\documentclass{article}
\usepackage{ijcai24}

\usepackage{times}
\usepackage{soul}
\usepackage{url}
\usepackage[hidelinks]{hyperref}
\usepackage[utf8]{inputenc}
\usepackage[small]{caption}
\usepackage{graphicx}
\usepackage{amsmath}
\usepackage{amsthm}
\usepackage{booktabs}
\usepackage{algorithm}
\usepackage{algorithmic}
\usepackage[switch]{lineno}
\usepackage{booktabs, multirow} 
\usepackage{soul}
\usepackage[table]{xcolor} 
\usepackage{changepage,threeparttable} 

\title{E-QGen: Educational Lecture Abstract-based Question Generation System}

\author{
Mao-Siang Chen\and
An-Zi Yen
\affiliations
Department of Computer Science, 
National Yang Ming Chiao Tung University, Taiwan\\
\emails
siang.cs09@nycu.edu.tw,
azyen@nycu.edu.tw
}

\begin{document}
\newcommand{\mathz}{%
  \text{\ooalign{\hidewidth\raisebox{0.2ex}{--}\hidewidth\cr$Z$\cr}}%
}
\maketitle

\begin{abstract}

To optimize the preparation process for educators in academic lectures and associated question-and-answer sessions, this paper presents E-QGen, a lecture abstract-based question generation system. 
Given a lecture abstract, E-QGen generates potential student inquiries. 
The questions suggested by our system are expected to not only facilitate teachers in preparing answers in advance but also enable them to supply additional resources when necessary.

\end{abstract}

\section{Introduction}

Teachers, when preparing their lecture content, are tasked with anticipating students' comprehension and any potential queries. 
Accordingly, to ensure the effectiveness of their instruction, they may adapt the course design and content to address these queries, for instance by preparing additional examples or allocating more time for explanation. 
In this case it would be helpful to have the use of a pedagogical support system capable of generating questions students might ask regarding lecture content.

Several studies generate questions based on specific text passages. 
Wikipedia is widely used as a resource based on which to generate questions~\cite{Liu2020AskingQT}.  
Moreover, several studies~\cite{zhou2017neural,yuan-etal-2017-machine,zhao-etal-2018-paragraph} utilize the SQuAD dataset~\cite{rajpurkar2016squad}, which consists of questions from Wikipedia paragraphs written by crowd workers.
As SQuAD was specifically created for machine reading comprehension, many of the answers to the questions are found within the paragraphs. 
As such, question generators trained on such a dataset may produce overly uniform questions,
often involving inquiries about the properties or descriptions of specific terms mentioned in the given article and predominantly feature questions starting with ``what''. 
The nature of such questions whose answers are readily found in the paragraphs may not accurately reflect the variety or depth of questions students tend to ask~\cite{ko2020inquisitive}.

Recently, large language models (LLMs) have demonstrated remarkable capabilities in natural language comprehension and generation, and have become foundational for natural language processing tasks across a wide range of domains~\cite{bommasani2021opportunities}.
Through prompt engineering, we instruct LLMs to generate any content that meets certain criteria or needs.
Nonetheless, LLMs employed to directly generate questions based on the provided lecture content may not fully meet the nuanced requirements of educational application scenarios,
because student inquiries often concern intricate details, or seek clarification of complex concepts, whereas LLM-generated questions typically focus more on general concepts of terminology in the lecture content.
This divergence may stem from the models' lack of context-specific understanding or training data that aligns with genuine student concerns. 

\begin{figure}[t]
    \centering
    \includegraphics[width=\linewidth]{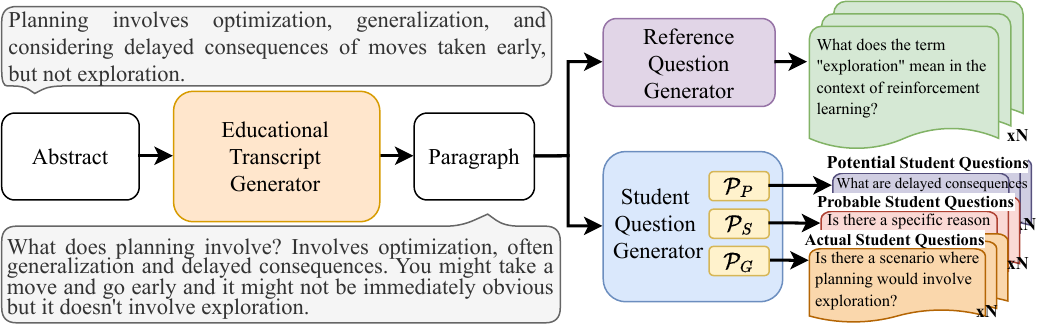}
    \caption{E-QGen system overview}
    \label{fig:overview}
\end{figure}

In this work, we construct E-QGen, a pilot system that produces questions that closely resemble those a student might ask, accurately reflecting their genuine learning needs and areas of confusion. 
To address the challenges associated with generating such questions, we propose a student question generator that uses multitask learning and LoRA~\cite{hu2021lora} fine-tuning. 
Figure~\ref{fig:overview} shows an overview of E-QGen.
E-QGen offers a service that allows teachers to input an abstract of the lecture content. 
Based on this abstract, the educational transcript generator automatically generates a complete lecture script, and the reference and student question generators subsequently produce suggested questions related to the generated script, aiding in comprehensive lesson planning. 
The reference question generator and student question generator are responsible for producing general questions and extended or in-depth questions, respectively. 
Our student question generator will produce three types of questions: actual student questions, probable student questions, and potential student questions.
Actual student questions closely align with what students typically ask; probable student questions reflect topics students may care about; and potential student questions estimate the inquiries students might have about course concepts, offering teachers high-quality suggestions without the need for costly APIs.

We construct a dataset by collecting real inquiries posed by students regarding the content of individual class sessions to fine-tune the model.
However, acquiring real-world student questions is challenging, primarily due to the inherent difficulty in obtaining a comprehensive collection of educational materials and their corresponding student-posed questions.
We collect publicly available lecture videos and transcripts uploaded by universities and research institutions to YouTube, along with user comments,
after which we extract the comments that contain inquiries specifically regarding the education transcripts.
Our contributions are threefold: 
\begin{itemize}
    \item We present a pioneer work on a pedagogically supported LLM that assists teachers in course preparation.  
    \item We construct a novel system to generate educational transcripts and associated student inquiries based on lecture abstracts.\footnote{E-QGen system: \url{https://nlplab.cs.nycu.edu.tw/demo}}\footnote{\url{https://youtu.be/SuiroLobtEU}}
    \item In addition to the demonstration system, we also construct a dataset comprising publicly available course transcripts and questions posed by students.\footnote{\url{https://github.com/NYCU-NLP-Lab/E-QGen}} 
\end{itemize}

\section{Dataset Construction}\label{sec:dataset}

\noindent \textbf{Educational Transcript and Comment Collection.}
Inspired by work~\cite{wang2023sight} that crawls user comments concerning educational transcripts, we collect both subtitles and comments from the MIT OpenCourseWare\footnote{\url{https://www.youtube.com/@mitocw}} and Stanford Online\footnote{\url{https://www.youtube.com/@stanfordonline}} YouTube channels.
These comments contain student thoughts, opinions, and inquiries.
The videos are sourced from 35 distinct playlists (courses), encompassing a total of 19,013 user comments from 963 videos. 
Note that the majority of comments express gratitude, which is consistent with statistics from SIGHT~\cite{wang2023sight}. 
Hence, we extract questions related to course content from the comments, and align those questions to the corresponding sections of the educational transcripts.

\noindent \textbf{Question Extraction.}
Given the large amount of comments, manual identification of questions would be labor-intensive and time-consuming. 
Therefore, we leverage the language understanding capabilities of LLMs as annotators to assist in question extraction by instructing them to label whether a comment is a question related to the course content. 
We employ three LLMs---\texttt{PaLM 2}~\cite{anil2023palm}, \texttt{GPT-3.5}~\cite{openai2023gpt35turbo}, and \texttt{GPT-4}~\cite{openai2023gpt4turbo}---for a labeling process that regards unanimously approved comments as questions.
Afterwards, we further manually verify that these comments are indeed questions about the course.
Comments that do not receive unanimous support are removed. 
This results in a total of 1,685 questions extracted from 19,013 comments.

\noindent \textbf{Paragraph Segmentation.}
Since the majority of videos exceed 40 minutes in duration, the subtitles are lengthy. 
To generate questions more precisely for individual concepts throughout the course content, we segment the transcripts into multiple paragraphs using the text tiling toolkit.\footnote{\url{https://www.nltk.org/_modules/nltk/tokenize/texttiling.html}} 
Important hyperparameters include $w$ and $k$, which represent the pseudosentence size and the number of sentences used in the block comparison method, respectively.
A smaller value of~$w$ leads to finer segmentation. 
For $k$, 10 is suggested as a proper value.
To ensure each paragraph conveys a complete and specific concept, $w$ and $k$ are set to 30 and 10, respectively.
We segment a total of 14,422 paragraphs.

\noindent \textbf{Question Alignment.}
As YouTube enables users to input timestamps in their comments to indicate the specific video segment their comment refers to, 419 questions can be mapped to a paragraph based on the timestamps.
For the 1,266 questions lacking timestamps, we propose question alignment using classification and an assessment of semantic relatedness.
For the classification method, we employ \texttt{PaLM 2} to determine whether a question is relevant to a specific paragraph. 
For the semantic relatedness assessment, we measure the cosine similarity between each paragraph and question. 
The paragraph and question representations are obtained through the \texttt{all\_minilm\_l6\_v2}~\cite{reimers2019sentence} and \texttt{PaLM 2} embedding models. 
For each question, we calculate the cosine similarity with each paragraph using two embeddings. 
Given these two sets of scores, we rank them separately, and take the top 10 matches from each ranking as potential question-paragraph pairs suggested by their respective embedding models.
These suggestions, combined with the classification results, are then submitted to a majority vote.
The question-paragraph pair that receives a majority of votes is used to identify the correct alignment.
In this way, we can align questions with multiple paragraphs.
The question-paragraph pairs identified based on user-typed timestamps and our alignment strategy are referred to as ``golden pairs'' and ``silver pairs'', respectively.
Golden pairs consist of actual questions accurately matched to specific paragraphs, and silver pairs, though derived from student questions, represent probable questions with less certainty regarding their specific paragraph alignment.
This results in 356 golden pairs and 4,434 silver pairs, with an average paragraph aligned with 1.2 and 2.3 questions, respectively.

\noindent \textbf{Data Augmentation.}
Due to limited question-paragraph pairs, we employ an LLM to augment our dataset and then use these to fine-tune our question generation model. 
Specifically, we instruct \texttt{GPT-4} to generate 20 potential student questions related to the course content based on the given transcript paragraphs.
We collect 4,829 question-paragraph pairs, which are referred to as ``platinum pairs''.

\section{E-QGen Implementation}
This section details the implementation of E-QGen, consisting of an educational transcript generator, followed by a student question generator and a reference question generator.

\noindent \textbf{Educational Transcript Generator.} 
To develop the educational transcript generator, we leverage the natural language comprehension and generation capabilities of LLMs, along with their extensive knowledge.

Simply by entering an abstract of the lecture content, our system automatically generates a paragraph that encapsulates the key aspects of that abstract. 

\noindent \textbf{Student Question Generator.} 
We adopt a multitask learning framework in which LoRA is applied to fine-tune the LLM. 
As mentioned in Section~\ref{sec:dataset}, since the number of golden pairs is limited, we postulate that fine-tuning with auxiliary tasks enhances the model's ability to generate high-quality questions that resemble those of students.
Specifically, we complete three distinct subtasks with specific prompts: (1)~$\mathcal{P}_G$ for generating actual student questions based on golden pairs $G$, (2)~$\mathcal{P}_S$ for formulating probable student questions from silver pairs $S$, and (3)~$\mathcal{P}_P$ for suggesting potential student questions associated with platinum pairs $P$.\footnote{All prompts are shown at \url{https://github.com/NYCU-NLP-Lab/E-QGen}.}

Formally, given an LLM parameterized by $\Phi$ and the task-specific parameter
increment $\Delta{\Phi}=\Delta{\Phi(\Theta)}$ encoded by a much smaller set of parameters $\Theta$, the fine-tuning process boils down to optimizing $\Theta$:
\begin{equation}
    \label{eq:lora}
    \max_{\Theta} \sum_{(x_i^k,y_i^k) \in \mathcal{D}^k }\sum_{t=1}^{|y|} (\log p_{{\Phi}_0 + \Delta\Phi(\Theta)}(y^k_{i,t} | x^k, y^k_{i,<t})),
\end{equation}
where $\mathcal{D}^k={\{(x^k_i, y^k_i)\}}_{i=1,\ldots,N}$ is the training data of subtask~$k$, 
$x_i^k$ is the $i$-th input generated by the task-specific prompt, and $y_i^k$ denotes the corresponding questions.
For instance, $x_i^G = \mathcal{P}_{G}(h_i)$ is the input of the first subtask to generate actual student questions $y_i^G$ for the $i$-th paragraph $h_i$.

\noindent \textbf{Reference Question Generator.}
Questions generated by the student question generator tend to delve deeper into specific concepts. 
To assist teachers in preparing a more comprehensive set of course questions, a reference question generator is constructed to produce general conceptual questions. 

\section{Experiment and Discussion}

\subsection{Experimental Setup}
We focus on evaluating the results of actual student question generation.
To assess whether the generated questions align most closely with those that students would pose, we ensure that questions from the same video are not included in both the training and test sets: 
this prevents the model from being exposed to similar questions during the training phase. 
The resulting question-paragraph pairs in the golden pairs are divided into training, validation, and test sets, with counts of 300, 10, and 46, respectively.
All silver and platinum pairs are used as training data.
\texttt{Vicuna-7b-v1.5}~\cite{zheng2023judging} and \texttt{GPT-3.5} are used as the base models for the student question generator and reference question generator, respectively. 
In our experiments, we set each generator in 
E-QGen to produce 20 questions for a given paragraph. 

\begin{table}[t]
  \centering
  \scriptsize
  \caption{Experimental results}
  \setlength\tabcolsep{3pt}
    \begin{tabular}{lcccc}
    \toprule
    Models & \multicolumn{1}{c}{ROUGE-1} & \multicolumn{1}{c}{ROUGE-2} & \multicolumn{1}{c}{ROUGE-L} & \multicolumn{1}{c}{BERTScore} \\
    \midrule
    GPT-3.5 & 0.2328 $\pm$ 0.0565 & 0.0589 $\pm$ 0.0202 & 0.1865 $\pm$ 0.0438 & 0.8619 $\pm$ 0.0074 \\
    GPT-4 & 0.2505 $\pm$ 0.0656 & 0.0658 $\pm$0.0210& 0.1967 $\pm$ 0.0507 & 0.8615 $\pm$ 0.0080\\
    SQ Only & 0.2418 $\pm$ \textbf{0.0660} & 0.0645 $\pm$ 0.0210 & 0.2019 $\pm$ \textbf{0.0528} & 0.8625 $\pm$ \textbf{0.1191} \\
    E-QGen & \textbf{0.2667} $\pm$ 0.0652 & \textbf{0.0866} $\pm$ \textbf{0.0238} & \textbf{0.2160} $\pm$ 0.0503 & \textbf{0.8642} $\pm$ 0.0857 \\
    \bottomrule
    \end{tabular}%
  \label{tab:main_exp}%
\end{table}%

\begin{table}[t]
  \centering
  \scriptsize
  \caption{Ablation results for student question generator}
    \begin{tabular}{lrrrr}
    \toprule
    Models & \multicolumn{1}{c}{ROUGE-1} & \multicolumn{1}{c}{ROUGE-2} & \multicolumn{1}{c}{ROUGE-L} & \multicolumn{1}{c}{BERTScore} \\
    \midrule
    SQ only & \textbf{0.2418} & \textbf{0.0645} & \textbf{0.2019} & \textbf{0.8625} \\
    \ w/o fine-tuning on $\mathcal{D}^S$ & 0.2245 & 0.0488 & 0.1905 & 0.8570 \\
    \ w/o fine-tuning on $\mathcal{D}^P$ & 0.2197 & 0.0455 & 0.1779 & 0.8578 \\
    \bottomrule
    \end{tabular}%
  \label{tab:ablation}%
\end{table}%

\subsection{Experimental Results}
Table~\ref{tab:main_exp} shows the results of the models that generate actual student questions for the given paragraph.
ROUGE-1, ROUGE-2, ROUGE-L~\cite{lin-2004-rouge}, and BERTScore~\cite{zhang2020bertscore} are adopted as the evaluation metrics.
Given that a paragraph may be aligned with multiple questions and the generators produce 20 questions for each paragraph, to evaluate the results of multiple candidates against multiple references, we compute the scores for all combinations of references and candidates across all metrics for each instance. 
For each reference, we select the candidate that achieves the highest average score in all metrics to calculate the overall score and standard deviation for that metric.
The scores before and after $\pm$ denote the overall score and the standard deviation.
Higher overall scores indicate better-quality questions generated by the model, whereas higher standard deviations suggest greater diversity in the questions produced.

``SQ Only'' denotes results generated using only the student question generator.
Experimental results show that ``SQ Only'' outperforms \texttt{GPT-3.5}.
In addition, E-QGen outperforms other LLMs, especially \texttt{GPT-4}.
This suggests that our multitask learning approach achieves better performance than models with more parameters.
Incorporating the reference question generator, the question generated by E-QGen not only aligns with relevant student inquiries but also provides a diverse range of questions for reference.

We also conducted an ablation study for the student question generator to investigate the impact when introducing different pairs of data.
In Table~\ref{tab:ablation}, the performance degrades most when excluding $\mathcal{D}^P$ in the training set.
This shows that pseudo-training data produced by the powerful LLM is effective for fine-tuning relatively small LLMs.

\section{Conclusion and Future Work}
This work demonstrates E-QGen, a pilot system that generates educational transcripts and questions that students are likely to ask, assisting teachers in proactively preparing course content.
To account for the limited number of question-paragraph pairs, we construct a dataset 
and implement a multitask learning framework to fine-tune the model.
Experimental results indicate that E-QGen achieves promising performance.
Moreover, the questions generated by our system surpass \texttt{GPT-4} both in similarity to student-posed questions and diversity.
At the current stage, our primary focus is on courses related to computer science.
In the future, we will broaden the applications of our method by extending it to cover courses across various fields.

\section*{Acknowledgement}
We thank the reviewers for their insightful comments.
This work was partially supported by the National Science and Technology Council, Taiwan, under grant NSTC 111-2222-E-A49-010-MY2, and by the Higher Education Sprout Project of the National Yang Ming Chiao Tung University and the Ministry of Education (MOE), Taiwan.

\bibliographystyle{named}
\bibliography{ijcai24}

\end{document}